\documentclass[runningheads]{llncs}

\usepackage{eccv}

\usepackage{eccvabbrv}
\usepackage[dvipsnames]{xcolor}
\usepackage{multirow}
\usepackage{multicol}
\usepackage{colortbl}
\usepackage{amsmath}
\usepackage{amssymb}
\usepackage{algorithm}
\usepackage{algpseudocode}
\usepackage{listings}
\definecolor{codegreen}{rgb}{0,0.5,0}
\definecolor{codeblue}{rgb}{0.25,0.5,0.5}
\definecolor{codegray}{rgb}{0.6,0.6,0.6}
\usepackage{pifont}
\newcommand{\cmark}{\ding{51}}%
\usepackage{bbm}
\usepackage{graphicx}
\usepackage{amsmath}
\usepackage{amssymb}
\usepackage{booktabs}
\usepackage{array}
\usepackage{multirow}
\usepackage{multicol}
\usepackage{makecell}
\usepackage[dvipsnames]{xcolor}

\usepackage{epsfig}
\usepackage{graphicx}
\usepackage{amsmath}
\usepackage{amssymb}
\usepackage{enumitem}

\usepackage{booktabs}
\usepackage{pifont}
\usepackage{bm}

\usepackage{caption}
\usepackage{subcaption}

\usepackage{float}
\usepackage{marvosym}
\usepackage[accsupp]{axessibility}  

\usepackage{hyperref}

\usepackage{orcidlink}

\begin{document}

\title{Learning High-resolution Vector Representation from Multi-Camera Images for 3D Object Detection} 

\titlerunning{VectorFormer}

\author{
Zhili Chen\inst{1\dagger}\orcidlink{0000-0002-8272-156X} \and
Shuangjie Xu\inst{1}\orcidlink{0000-0003-0150-7068} \and
Maosheng Ye\inst{1}\orcidlink{0000-0001-8470-685X} \and
Zian Qian\inst{1}\orcidlink{0000-0001-5147-9689} \and
Xiaoyi Zou\inst{2}\orcidlink{0000-0003-0074-1135} \and
Dit-Yan Yeung\inst{1}\orcidlink{0000-0003-3716-8125} \and
Qifeng Chen\inst{1}\textsuperscript{\Letter}\orcidlink{0000-0003-2199-3948}
}

\authorrunning{Z.~Chen et al.}

\institute{
$^1$HKUST \qquad
$^2$DeepRoute.AI \\
\email{\{zchenei, shuangjie.xu, myeag, zqianaa\}@connect.ust.hk}, \email{xiaoyizou@deeproute.ai}, \email{\{dyyeung, cqf\}@cse.ust.hk}
}

\renewcommand{\thefootnote}{}
\footnotetext{$^\dagger$Work done during an internship at DeepRoute.AI.\\
\textsuperscript{\Letter}Corresponding author.}

\maketitle

\begin{abstract}
The Bird's-Eye-View (BEV) representation is a critical factor that directly impacts the 3D object detection performance, but the traditional BEV grid representation induces quadratic computational cost as the spatial resolution grows. To address this limitation, we present a new camera-based 3D object detector with high-resolution vector representation: VectorFormer. The presented high-resolution vector representation is combined with the lower-resolution BEV representation to efficiently exploit 3D geometry from multi-camera images at a high resolution through our two novel modules: vector scattering and gathering. To this end, the learned vector representation with richer scene contexts can serve as the decoding query for final predictions. We conduct extensive experiments on the nuScenes dataset and demonstrate state-of-the-art performance in NDS and inference time. Furthermore, we investigate query-BEV-based methods incorporated with our proposed vector representation and observe a consistent performance improvement. Project page at \url{https://github.com/zlichen/VectorFormer}.
\end{abstract}    
\section{Introduction}
\label{sec:intro}
Integrating multi-camera information into a powerful, unified representation has been a challenging task in the perception system for autonomous driving cars and robotics~\cite{li2022delving, ma2022vision}. To better fuse features from different cameras with various viewpoints, BEV (bird's-eye-view) is proposed, which constructs a unified discrete spatial space in which features are transformed through forward projection~\cite{philion2020lift, huang2021bevdet} or backward projection~\cite{li2022bevformer,li2023fb}. 

\begin{figure}
     \centering
         \begin{subfigure}[t]{0.22\textwidth}
             \centering
             \includegraphics[width=\textwidth]{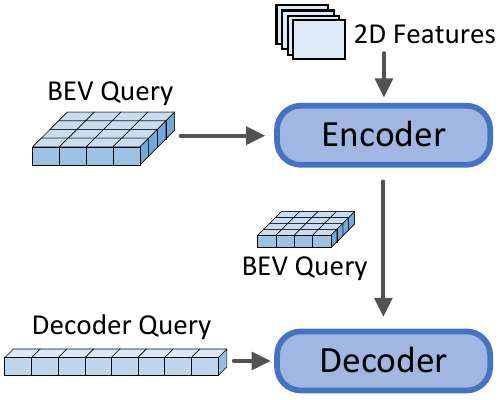}
             \caption{The Query-BEV Architecture}
             \label{fig:CoreA}
         \end{subfigure}
         \begin{subfigure}[t]{0.24\textwidth}
             \centering
             \includegraphics[width=\textwidth]{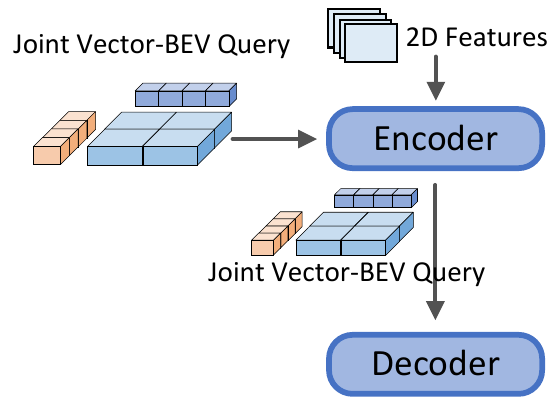}
             \caption{Our proposed VectorFormer}
             \label{fig:CoreB}
         \end{subfigure}
         \begin{subfigure}[t]{0.22\textwidth}
             \centering
             \includegraphics[width=\textwidth]{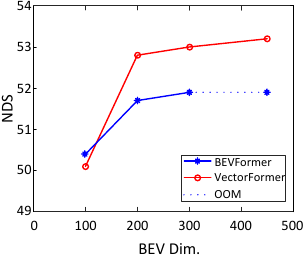}
             \caption{Resolution \& NDS}
             \label{fig:motivation_nds}
         \end{subfigure}
         \begin{subfigure}[t]{0.245\textwidth}
             \centering
             \includegraphics[width=\textwidth]{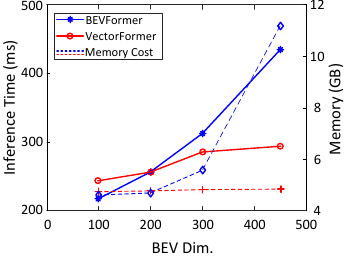}
             \caption{Resolution \& Resource Cost}
             \label{fig:motivation_time}
         \end{subfigure}

        \caption{\textbf{Comparison between the typical query-BEV architecture~\cite{li2022bevformer} and our proposed VectorFormer with the novel vector query.} Compared to the traditional design in Fig.~\ref{fig:CoreA}, our vector queries in Fig.~\ref{fig:CoreB} are encoded with finer-grained scene contexts, which transform into more accurate 3D predictions by decoder. Fig.~\ref{fig:motivation_nds} and Fig.~\ref{fig:motivation_time} are the effectiveness and efficiency comparisons. Compared to the architecture in Fig.~\ref{fig:CoreA}, the performance (Fig.~\ref{fig:motivation_nds}) of VectorFormer keeps benefitting by scaling up the representation resolution without leading to a long inference time and high memory cost (Fig.~\ref{fig:motivation_time}). Noted that BEVFormer will be out-of-memory (OOM) when training under the BEV resolution of $450\times 450$ with NVIDIA A100 40GB GPUs.}
        \label{fig:motivation}
\end{figure}

One mainstream approach to lifting multi-camera features into BEV space is through backward projection methods, with BEVFormer~\cite{li2022bevformer} pioneering this technique by utilizing transformers and deformable attention to sample camera features for each BEV location.
Works of ~\cite{huang2021bevdet,Wang_2023_ICCV,yang2023bevformer,liu2023vision} follow a similar pipeline while replacing with polar representations or more efficient operations including sparse queries. To sum up, the standard BEV formulation process could be two steps: 1) construct BEV grids and associate them with image features according to the projection matrix. 2) Update the feature of each grid in the BEV space through learnable grid sampling based on the constructed association of the previous step~\cite{li2022bevformer}. However, we observe an apparent dilemma in the BEV representation. The BEV grid with higher spatial resolution can lead to a decent perception performance, as illustrated in Fig.~\ref{fig:motivation_nds}. It is primarily due to the fact that a more fine-grained BEV grid results in a higher sampling frequency in world coordinates, which ensures a more precise association with the image features~\cite{li2022delving}. As a trade-off, the computational cost (Fig.~\ref{fig:motivation_time} and Tab.~\ref{tab:resolution}) of the BEV formulation grows quadratically with the BEV grid resolution because of the denser sampling. To solve the overhead of BEV representation formulation, a series of works~\cite{chen2022efficient,zhou2022cross,huang2022bevpoolv2} design efficient operations for view transformations through hashing the corresponding position calculation, which has predominantly focused on the input resolution and optimized the 2D-to-3D projection. However, they have not addressed the optimization of the BEV resolution itself.

Different from the prior approaches to solving this problem, we propose our VectorFormer, as shown in Fig.~\ref{fig:CoreB}. It is based on vector factorization on the crucial regions into two low-rank tensor components, which consists of a pair of learnable vector queries that represent the $x$-axis and $y$-axis to compress those scenario representations at a finer granularity. As the first time proposing vector factorization by vector queries in query-BEV-based methods, we can encode crucial regions at higher BEV resolution into the compact vector queries and further decode into the 3D predictions.
Consequently, our model achieves superior performance even when operating under comparable resolutions in Fig.~\ref{fig:motivation_nds} (e.g., BEVFormer with BEV size $200 \times 200$ v.s. ours with BEV size $150 \times 150$ and vector tensor size $200$).
Furthermore, we represent the innovative sparse high-resolution (HR) BEV features for the critical regions through our vector query scattering module and then interact with the multi-view image features, achieving a substantial reduction in both time and memory complexity. This stands in stark contrast to the $O(n^2)$ complexity observed in previous BEV methods, as our vector representation introduces a more efficient $O(n)$ complexity. This significant reduction contributes directly to the overall improvement in both inference time and memory efficiency within our proposed framework, as shown in Fig.~\ref{fig:motivation_time}.

Our contributions are summarized as follows:

\begin{itemize}
 \renewcommand{\labelitemi}{\textbullet}
    \item We introduce the VectorFormer, which incorporates a novel Vector representation, a first for query-BEV-based methods. Our approach leverages the factorization philosophy which enables efficient modeling of spatial and temporal at a higher resolution. It consistently achieves performance gains as the resolution increases.

    \item We design the Vector Query Scattering and the Vector Query Gathering modules to learn the expressive high-resolution Vector queries from images. 
    Thanks to the proposed vector representation, our approach benefits from fine BEV granularity with a more efficient $O(n)$ complexity, making additional overhead negligible.

    \item We conduct extensive experiments on the challenging nuScenes \cite{caesar2020nuscenes} and Waymo \cite{sun2020scalability} datasets. VectorFormer outperforms state-of-the-art camera-based 3D object detectors and is comparable to leading depth-aware methods.
    
\end{itemize}
\section{Related Work}
\subsection{Sparse Query 3D detector}
As transformer-based methods like DETR~\cite{carion2020end} generate great performance gain in 2D object detection~\cite{wang2021pnp, carion2020end, roh2021sparse, zhu2020deformable, li2022dn,jia2023detrs}, researchers begin to explore its potential in 3D object detection~\cite{wang2022detr3d, luo2022detr4d,lin2022sparse4d,lin2023sparse4dv2,lin2023sparse4dv3,liu2023sparsebev,yao2023dynamicbev,liu2022petr,liu2022petrv2,Wang_2023_ICCV,shu20233dppe}. As the pioneer of extending DETR to 3D object detection, DETR3D~\cite{wang2022detr3d} projects the 3D object center back to 2D multi-view images and iteratively updates the object query from the corresponding sampled image features. Although DETR3D is very efficient, the limitation still exists. The inaccurate object location predicted previously will cause error accumulation in the image feature sampling process. Besides, the model has less information on the global features. To solve the above limitations, PETR~\cite{liu2022petr} and PETRv2~\cite{liu2022petr} transfer 2D multi-view image features into 3D perception features through 3D position embedding. Sparse4D~\cite{lin2022sparse4d} and StreamPETR~\cite{Wang_2023_ICCV} aggregating temporal information to help the detection of the current object query. Although sparse query 3D detectors have a simple structure and are efficient, they can still hardly reach the state-of-the-art performance, while our method can reach the state-of-the-art performance while only inducing negligible additional overhead in computational cost.

\subsection{Dense BEV Feature 3D Detector}
The dense bird's-eye view (BEV) feature 3D detector has drawn great attention in recent years due to its strong performance and wide application in autonomous driving~\cite{li2022bevformer, yang2023bevformer, philion2020lift, huang2021bevdet, huang2022bevdet4d, li2023bevdepth, li2023bevstereo, qi2023ocbev, wu2023heightformer, li2023dfa3d,li2024bevnext}. Different from sparse query detectors, dense BEV feature detectors tend to build an explicit BEV feature representation from multi-view 2D images and further utilize it for other perception tasks. BEVFormer~\cite{li2022bevformer} and BEVFormerv2~\cite{yang2023bevformer} build a BEV query and utilize spatial cross-attention to update the BEV query from the 2D multi-view images with the sampled points in 3D space. Besides, they also propose a temporal self-attention to fuse the temporal features. Based on BEVformer, OCBEV~\cite{qi2023ocbev} has an object-aligned temporal fusion module to align the fast-moving objects, and a heatmap to provide prior knowledge of the position of the objects in the decoding stage. Unlike BEVFormer, BEVDet~\cite{huang2021bevdet} and BEVDet4D~\cite{huang2022bevdet4d} attempt to build BEV features in a forward way. They first extract the features from multi-view 2D images and then transform them into BEV with the help of the predicted depth and further encode them into BEV features.

Although BEV feature representation has rich information, the computation cost of building the query from multi-view 2D images is also expensive. The computational cost will grow quadratically as the resolution of the BEV query increases, therefore limiting the information contained within the BEV features. Our method can handle the BEV query with a high resolution while inducing negligible additional computational cost.

\section{Method}
\label{method}
\begin{figure}[t!]
\centering
\includegraphics[width=1.0\textwidth]{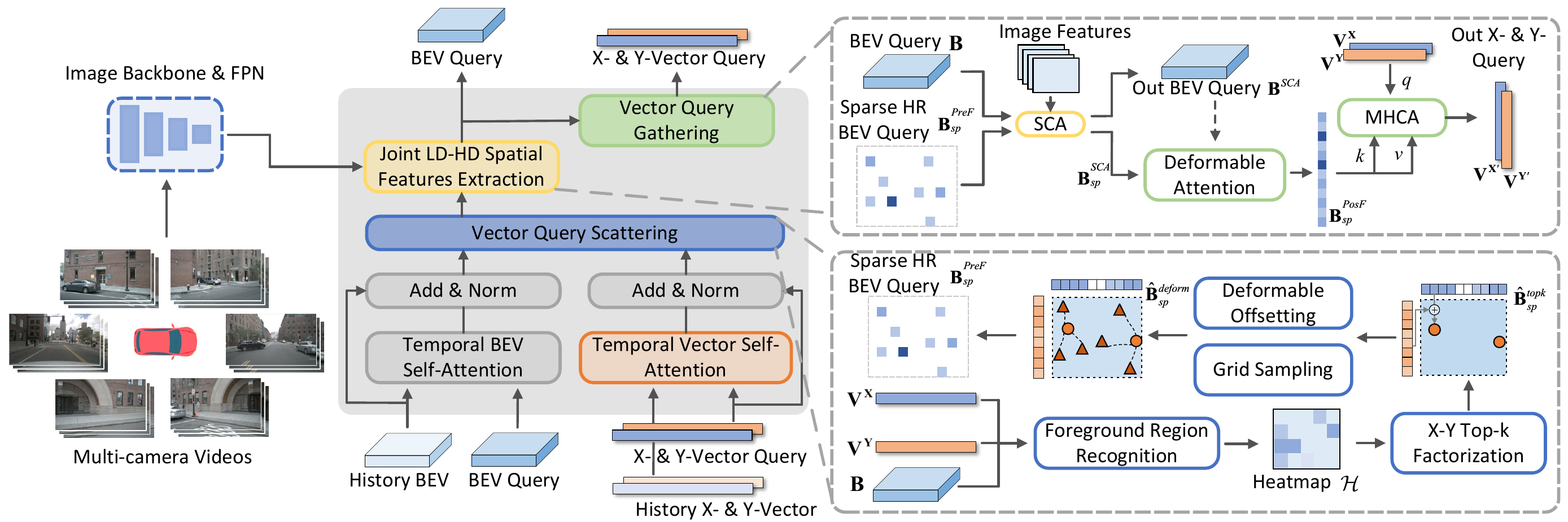}

\caption{\textbf{The overall framework of our proposed VectorFormer.} 
In the middle, the encoder takes the BEV query and Vector query at the current timestamp and the ones from the previous timestamp as inputs for interacting with multi-view image features at each layer. The learned representations are further transformed into 3D predictions by the decoder. We highlight our innovations for learning high-resolution vector representation within the encoder and zoom in on the designs of Vector Query Scattering (Fig.~\ref{fig:scatter} for the details) and the Vector Query Gathering at the right. In the Vector Query Scattering module, we represent the sparse high-resolution (HR) BEV features by first recognizing the foreground regions and then compositing them with the factorized Vector queries of $\mathbf{V^X}$ and $\mathbf{V^Y}$. After jointly interacting with the image features, we finally factorize the learned HR BEV features back into $\mathbf{V^X}$ and $\mathbf{V^Y}$ through the Vector Query Gathering module.}
\label{fig:overall}
\end{figure}

We demonstrate the overall architecture of VectorFormer in Fig.~\ref{fig:overall}. Inheriting the genre of query-BEV 3D detectors~\cite{li2022bevformer, yang2023bevformer}, it consists of the image backbone, encoder, and decoder head. We mainly focus on innovating the encoder and the decoder head for high-resolution (HR) Vector representation learning. We introduce the HR BEV features factorization by two low-rank Vector queries in Sec.~\ref{sec3.1}. In Sec.~\ref{sec3.2}, we demonstrate how to locate the sparse informative regions and further guide the HR BEV features composition for those regions by the Vector queries. Later, we present how we interact the LR and sparse HR BEV features with the multi-view image features in a unified manner at Sec.~\ref{sec3.3}. In Sec.~\ref{sec3.4}, we update the Vector queries by gathering information from the learned HR BEV features with the multi-head cross-attention mechanism~\cite{vaswani2017attention} (\texttt{MHCA}). By further enhancing through the temporal modeling (Sec.~\ref{sec3.5}), the Vector queries are considered superior decoding queries and further transform into accurate predictions in the decoder.

\subsection{Vector Query}
\label{sec3.1}
In the real world, the objects have a much smaller scale compared to the whole perception range~\cite{caesar2020nuscenes,fan2022embracing}. We argue that not all regions are equally important, and we instead overcome the limited computational resources by learning the BEV features at high resolution for a sparse set of crucial regions. Inspired by the factorization philosophy in the field of volume rendering~\cite{chen2022tensorf,fridovich2023k,huang2023tri}, we propose to factorize the sparse HR BEV features by two vector-shaped queries of $ \mathbf{V^X}\in\mathbb{R}^{W_{HR}\times C}$ and $ \mathbf{V^Y}\in \mathbb{R}^{H_{HR}\times C}$. Additionally, we initialize their accompanying learnable positional embeddings, $\mathbf{PE^X}$ and $\mathbf{PE^Y}$, having the same dimensionality as $\mathbf{V^X}$ and $\mathbf{V^Y}$.

With two low-rank tensors $\mathbf{V^X}$ and $\mathbf{V^Y}$, the creation of a sparse HR BEV feature is achieved to enhance granularity. Our work introduces three distinct BEV representations:
1) Traditional BEV: Denoted as $\mathbf{B}\in \mathbb{R}^{H_{LR}\times W_{LR}\times C}$, this corresponds to the low-resolution full-grid BEV query.
2) Sparse BEV: Comprising sparse sets of the LR and HR BEV features, denoted as $\mathbf{B}{sp}\in \mathbb{R}^{N_{LR} \times C}$ and $\mathbf{\hat{B}}{sp}\in \mathbb{R}^{N_{HR}\times C}$, respectively. The latter enriches contextual features with minimal additional computational overhead.

\noindent\textbf{Data Representation} The factorization operates on:
\begin{itemize}[label=-]
\item Vector query representation $\mathbf{V^X} =(\mathbf{c^x},\mathbf{v^x})$, where $\mathbf{c^x} \in \mathbb{R}^{W_{HR} \times 2}$ denotes the 2D center coordinate set of all vector cells, which is represented as $(x, 0)$ uniformly located along the $x$-axis vector representation. $\mathbf{v^x} \in \mathbb{R}^{W_{HR} \times C}$ denotes the feature set of all vector cells. Similar definition could be derived for $\mathbf{V}^{\mathbf{Y}}$, $\mathbf{PE}^\mathbf{X}$ and $\mathbf{PE}^\mathbf{Y}$.

\item Sparse HR BEV query representation is defined as $\mathbf{\hat{B}} =(\mathbf{c},\mathbf{b})$, where $\mathbf{c}$ is the sparse 2D coordinate set and $\mathbf{b}$ are their corresponding composited HR BEV features. We utilize an operation $f(\cdot)$ to obtain the sparse HR BEV queries which takes the $\mathbf{V^X}$, $\mathbf{V^Y}$, and $\mathbf{c}=[\mathbf{c^x}.\mathbf{x}, \mathbf{c^y}.\mathbf{y}]$ as input:
\begin{equation}
\label{eqn:1}
\begin{aligned}
\mathbf{\hat{B}}_{sp} = f(\mathbf{V^X}&, \mathbf{V^Y}, \mathbf{c}), \\
\mathbf{b} =\texttt{GridSample}(\mathbf{V^X}, \mathbf{c^x}) &+ \texttt{GridSample}(\mathbf{V^Y}, \mathbf{c^y}),
\end{aligned}
\end{equation}
where we utilize an operation $\texttt{GridSample}(\mathbf{F}, \mathbf{C})$~\cite{jaderberg2015spatial} to bilinearly sample features by taking features $\mathbf{F}$ and the 2D coordinates $\mathbf{C}$ as input.

\end{itemize}

\begin{figure}[t!]
\centering
\includegraphics[width=1.0\textwidth]{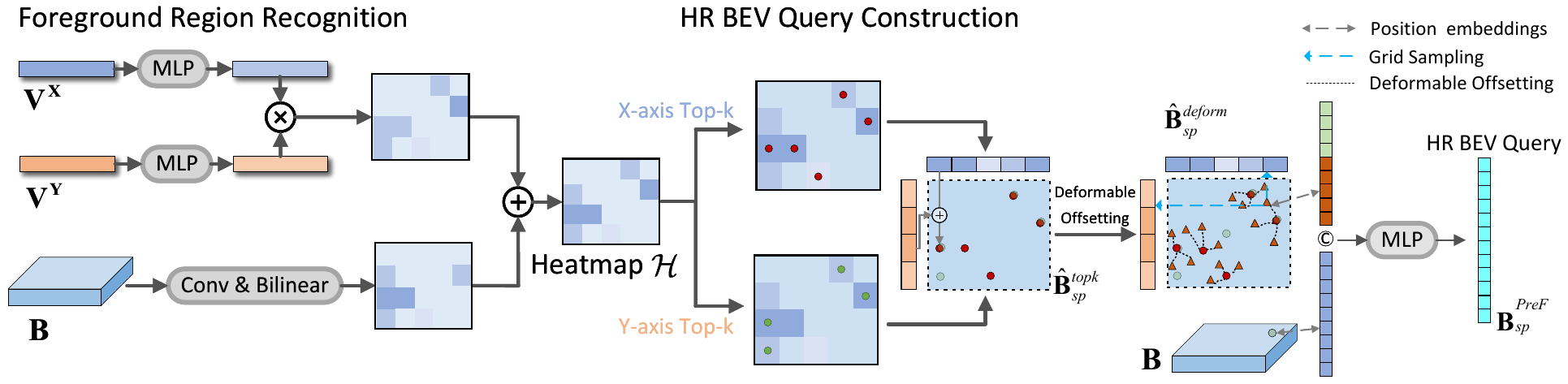}

\caption{\textbf{The design details for the Vector Query Scatter Module of our proposed VectorFormer.} We recognize the foreground regions by first predicting an objectness heatmap using high-resolution (HR) Vector queries $\mathbf{V^X}$ and $\mathbf{V^Y}$ and low-resolution (LR) BEV queries as input and then locate the foreground regions by taking the directional Top-$k$ ($k=1$ here as demonstration) on the heatmap. Next, we apply deformable offsetting for learning to cover more informative regions. We finally construct the sparse HR BEV by fusing the factorized Vector queries and the grid sampled LR BEV features for these informative regions.}
\label{fig:scatter}
\end{figure}

\subsection{Vector Query Scattering}
\label{sec3.2}
Throughout the BEV encoding module, the Vector queries $\mathbf{V^X}$ and $\mathbf{V^Y}$ are learned to encode with the finer-grained foreground contexts. The LR BEV queries are uniformly sampled, which are learned with more background contexts. As shown in Fig.~\ref{fig:scatter}, the Vector Query Scattering module needs first to utilize a heatmap to recognize the critical foreground areas and eventually composite a sparse set of finer-grained BEV queries for the important regions deformed from the proposed areas indicated by the heatmap. These finer-grained BEV queries will exploit the spatial geometry of these important regions from the image features.

\noindent
\textbf{Foreground Region Recognition}
To predict the grid-shape heatmap from the Vector queries, which indicates the probability of the objectness, we first apply an MLP with two hidden layers to the Vector queries. Then, we conduct matrix multiplication between $\mathbf{V^X}$ and $\mathbf{V^Y}$ resulting in the heatmap $\mathbf{h} \in \mathbb{R}^{H_{HR}\times W_{HR}}$. We apply a lightweight convolution module with two layers to the LR BEV feature map, resulting in another heatmap $\mathbf{h}^\prime$. The final heatmap $\mathcal{H}$ is formed by adding the $\mathbf{h}$ with the bilinear upsampled heatmap of $\mathbf{h}^\prime$:
\begin{equation}
\label{eqn:2}
\begin{aligned}
\mathbf{h} = \texttt{MLP}_1 & (\mathbf{V^X}) \texttt{MLP}_2 (\mathbf{V^Y})^T, \mathbf{h}^\prime = \texttt{Conv}(\mathbf{B}),\\
& \mathcal{H} = \mathbf{h} + \texttt{Upsample}(\mathbf{h}^\prime),
\end{aligned}
\end{equation}
where the $\texttt{Conv}$ module transforms the BEV query $\mathbf{B}$ by $256\rightarrow 64\rightarrow 1$, followed by a bilinear upsampling layer denoted as $\texttt{Upsample}$. We use a gaussian focal loss~\cite{lin2017focal,zhou2019objects,law2018cornernet,huang2021bevdet} to supervise the the heatmap predictions, resulting in $\mathcal{L}_{hm} = \mathcal{L}_{focal}(\mathcal{H})+ \mathcal{L}_{focal}(\mathbf{h}^\prime)$.

\noindent
\textbf{HR BEV Query Construction}
We select the foreground regions along $x$ and $y$ directions separately instead of locating them globally. This practice allows us to derive a uniform number of sparse HR BEV queries for each vector cell of $\mathbf{V^X}$ and $\mathbf{V^Y}$, which can further enable us to parallelly aggregate the learned sparse HR BEV queries into the vector queries via Vector Query Gathering module, as discussed in Sec.~\ref{sec3.4}. Specifically, we take the predicted heatmap as input and then apply the Top-$k$ operation along the $H$ and $W$ dimensions, as illustrated in Fig.~\ref{fig:scatter}. We obtain $k$ proposal positions for each entry of both vector queries, resulting in a sparse set of 2D coordinates $\mathbf{c}_{topk}$ with a total number of $(W_{HR}+H_{HR})*k$. The sparse HR BEV queries $\mathbf{\hat{B}}_{sp}^{topk}\in\mathcal{R}^{(W_{HR}+H_{HR})*k*C}$ for these proposal positions are constructed as defined in Eqn.~\ref{eqn:1}:
\begin{equation}
\label{eqn:3}
\begin{aligned}
\mathbf{\hat{B}}_{sp}^{topk} &= f(\mathbf{V^X}, \mathbf{V^Y}, \mathbf{c}_{topk}).
\end{aligned}
\end{equation}

However, the directional selection strategy might result in constructing sparse HR BEV queries located in the less important areas, especially for the ones at the margin of the perception range. To further exert the expressiveness for their representing local areas, we deform the proposal positions to more informative regions, namely deformable offsetting. Inspired by~\cite{zhu2020deformable,dai2017deformable}, we predict $\delta$ offsets with an $\texttt{MLP}$ for each proposal position by taking the previously constructed sparse HR BEV queries as input.

As a result, we ground each vector cell on Vector queries with $k * \delta$ of 2D positions on the BEV grid defined at a higher resolution. We obtain the coordinate set $\mathbf{c}_{deform}$ has a total number of $(W_{HR} + H_{HR}) * k * \delta$ after deformable offsetting. The sparse HR BEV queries $\mathbf{\hat{B}}_{sp}^{deform}$ with a dimensionality of $\mathbb{R}^{(W_{HR}+H_{HR})*k*\delta*C}$ is obtained as Eqn.~\ref{eqn:1}:
\begin{equation}
\label{eqn:4}
\begin{aligned}
\mathbf{\hat{B}}_{sp}^{deform} &= f(\mathbf{V^X}, \mathbf{V^Y}, \mathbf{c}_{deform}).
\end{aligned}
\end{equation}

However, approximating the BEV with the vector representation only would definitely bring information loss. Therefore, we jointly interact the LR BEV query $\mathbf{B}$ and the sparse HR BEV query $\hat{\mathbf{B}}_{sp}^{deform}$ with the image features. To extract the complement information from the image features, we additionally let the sparse HR BEV queries aware of their corresponding LR BEV queries. Specifically, we first obtain their spatially aligned sparse LR BEV queries through the operation of $\texttt{GridSample}$. Finally, we fuse the features through an MLP, which takes the concatenation of the sparse LR and HR BEV queries as input:
\begin{gather}
\label{eqn:5}
\mathbf{B}_{sp} = \texttt{GridSample}(\mathbf{B}, \mathbf{c}_{deform}), \\
\label{eqn:6}
\hat{\mathbf{B}}^{PreF}_{sp} = \texttt{MLP}([\mathbf{B}_{sp}, \hat{\mathbf{B}}_{sp}^{deform}]),
\end{gather}
where $\hat{\mathbf{B}}^{PreF}_{sp}$ are the resulting pre-fused sparse HR BEV queries before interacting with the image features.

\subsection{Joint LR-HR Spatial Features Extraction}
\label{sec3.3}
The LR and the sparse HR BEV queries exploit the 3D world geometric from image features through deformable-attention~\cite{zhu2020deformable} according to the sampled 3D-2D mapping reference points. As discussed in Sec.~\ref{sec3.2}, the sparse HR BEV queries $\hat{\mathbf{B}}_{sp}$ are derived from a finer-grained BEV grid, which has a more precise 3D-2D coordinates mapping. Compared to uniformly defined LR BEV queries ${\mathbf{B}}$, they focus on learning finer-grained BEV features for the foreground objects. For efficiency, we use a shared spatial cross-attention (SCA) module from BEVFormer~\cite{li2022bevformer} to jointly interact with the image features to exploit the spatial geometric. The SCA module takes the BEV queries at different resolution granularity as attention queries, and the multi-view image features are considered keys and values:
\begin{equation}
\label{eqn:7}
\begin{aligned}
{[}\mathbf{B}^{SCA},\hat{\mathbf{B}}^{SCA}_{sp}] &= \texttt{SCA} \big(  [\mathbf{B},\hat{\mathbf{B}}^{PreF}_{sp}], \mathbf{F}_t, \mathcal{P}[p_{LR}, p_{HR}]) \big),\\
\end{aligned}
\end{equation}
where $[ \boldsymbol{\cdot} ]$ denotes the operation of concatenation, $\mathbf{B}$ and $\hat{\mathbf{B}}^{PreF}_{sp}$ are LR BEV queries and sparse HR BEV queries, $\mathbf{F}_t$ is the multi-view image features are used as keys and values, $\mathcal{P}$ is the 3D-2D projection function, $p_{LR}$ and $p_{HR}$ are the 3D coordinates of the reference points correspond to the LR and the sparse HR BEV queries, respectively.

The LR and sparse HR BEV queries independently interact with the image features to extract complement geometric features. Compared to the sparse HR BEV queries, LR BEV queries are evenly defined in the 3D world, which contains global environmental semantics. On the other hand, the sparse HR BEV queries are more foreground-focused and represent these regions at a finer-grained resolution. Before factorizing these sparse HR BEV queries into the Vector queries of $\mathbf{V^X}$ and $\mathbf{V^Y}$, we enhance the sparse HR BEV queries with more scene contexts by applying a deformable attention~\cite{zhu2020deformable} to the LR BEV queries:
\begin{equation}
\label{eqn:8}
\begin{aligned}
\hat{\mathbf{B}}^{PosF}_{sp}  = \texttt{Deform}(\hat{\mathbf{B}}^{SCA}_{sp}, \mathbf{B}^{SCA}, \mathbf{c}_{deform}) + \hat{\mathbf{B}}^{SCA}_{sp},
\end{aligned}
\end{equation}
where the sparse HR BEV features $\hat{\mathbf{B}}^{SCA}_{sp}$ regarded as the attention queries, the LR BEV features $\mathbf{B}^{SCA}$ as keys and values. The extracted contexts feature from LR BEV features have a skip connection to $\hat{\mathbf{B}}^{SCA}_{sp}$, eventually output with the post-fused $\hat{\mathbf{B}}^{PosF}_{sp}$.

\subsection{Vector Query Gathering}
\label{sec3.4}
With the learned sparse HR BEV queries from the image features, we further aggregate them into vector queries through Vector Query Gathering. Benefiting from the evenly derived $k * \delta$ sparse HR queries for each vector cell, we can utilize the multi-head cross-attention mechanism $\texttt{MHCA}$~\cite{vaswani2017attention} to query information from the learned HR queries back to both of the vector queries $\mathbf{V^X}$ and $\mathbf{V^Y}$.

For the aggregation of $\mathbf{V^X}$, it absorbs the first $N^{x}_{HR}=W_{HR} * k * \delta$ of the sparse HR BEV queries which are derived from each cell of $\mathbf{V^X}$ along the side of the $y$-axis. And the remaining $N^{y}_{HR}=H_{HR} * k * \delta$ of those will squeeze into $\mathbf{V^Y}$. To remedy the information loss of the dimension about to collapse while conducting vector query gathering, we append the vector positional embeddings of each other to the sparse HR BEV queries, which are obtained by: 
\begin{equation}
\label{eqn:9}
\begin{aligned}
\mathbf{pe}^\mathbf{x} &= \texttt{GridSample}(\mathbf{PE^X}, \mathbf{c}_{deform}),\\
\mathbf{pe}^\mathbf{y} &= \texttt{GridSample}(\mathbf{PE^Y}, \mathbf{c}_{deform}),\\
\mathbf{pe}_{sp} &= \big[\mathbf{pe}^{\mathbf{y}}_{[0:N^{x}_{HR}]}, \mathbf{pe}^{\mathbf{x}}_{[N^{x}_{HR}:(N^{x}_{HR}+N^{y}_{HR})]}\big],
\end{aligned}
\end{equation}
where we slice the first $N^{x}_{HR}$ positional embeddings from $\mathbf{pe}^{\mathbf{y}}$ for the sparse HR BEV query about to aggregate into $\mathbf{V^{X}}$. The following $N^{y}_{HR}$ HR BEV queries that will go to $\mathbf{V^{Y}}$ uses the corresponding positional embeddings from $\mathbf{pe}^{\mathbf{x}}$.

The Vector Query Gathering process can be formulated as follows:
\begin{equation}
\label{eqn:10}
\begin{aligned}
&[\mathbf{V^X}^\prime,\mathbf{V^Y}^\prime] = \texttt{MHCA}(q, k, v, q_{pos}, k_{pos}), \\
&q = [\mathbf{V^X},\mathbf{V^Y}], q_{pos} = [\mathbf{PE^X}, \mathbf{PE^Y}], \\
&k = v = \hat{\mathbf{B}}^{PosF}_{sp}, k_{pos} = \mathbf{pe}_{sp},
\end{aligned}
\end{equation}
where the attention query $q$ is the concatenation of layer input $\mathbf{V^X}$ and $\mathbf{V^Y}$, and their corresponding positional embeddings $\mathbf{PE^X}$ and $\mathbf{PE^Y}$ are regarded as the query position embeddings $q_{pos}$. The sparse HR BEV features $\hat{\mathbf{B}}^{PosF}_{sp}$ are considered attention keys and values. The key position embeddings are $\mathbf{pe}_{sp}$. We then obtain the resulting $\mathbf{V^X}^\prime$ and $\mathbf{V^Y}^\prime$, which will be passed to the following encoding layer for further refinements.

\subsection{Model Architecture Details}
\label{sec3.5}
\noindent
\textbf{Temporal Features Extraction} helps to reason about the existence of highly occluded objects and infer the motion of the objects. Similar to the BEV representation~\cite{li2022bevformer}, the encoded Vector query from the previous frame can also be considered as strong priors, which could further improve the scene understanding ability. We apply multi-head self-attention~\cite{vaswani2017attention} (\texttt{MHSA}) among vector queries of the previous frame and those of the current frame and use the average operation to fuse the attention output.

\noindent
\textbf{Vector Queries As Decoding Queries}
The Vector queries are encoded with strong spatial and temporal contexts. Different from the traditional practices~\cite{zhu2020deformable,li2022bevformer,yang2023bevformer} that use randomly initialized learnable embeddings as the decoder queries, we use our learned Vector queries as the decoding query to enhance the decoding head. The total number of the decoding queries equals the summation length of the vector queries representing the $x$-axis and $y$-axis with the number of $H_{HR} + W_{HR}$. Inspired by~\cite{jia2023detrs}, the vector queries at intermediate encoder layers are fed into the decoder and supervised to make 3D predictions for network training acceleration.

\begin{table}[t!]
\begin{center}
\centering
\resizebox{\textwidth}{!}{
\setlength{\tabcolsep}{0.3em}
\renewcommand{\arraystretch}{1.05}
\begin{tabular}{l | l | c c | c c c c c}
\Xhline{2.0\arrayrulewidth}
Method & Backbone & NDS~$\uparrow$ & mAP~$\uparrow$ & mATE~$\downarrow$ & mASE~$\downarrow$ & mAOE~$\downarrow$ & mAVE~$\downarrow$ & mAAE~$\downarrow$ \\
\Xhline{2.0\arrayrulewidth}

BEVDet$\dagger$~\cite{huang2021bevdet} & V2-99~\cite{lee2019energy} & 48.8 & 42.4 & 52.4 & 24.2 & 37.3 & 95.0 & 14.8 \\

BEVDet4D$\dagger$~\cite{huang2022bevdet4d} & Swin-B~\cite{liu2021swin} & 56.9 & 45.1 & \textbf{51.1} & \textbf{24.1} & 38.6 & \textbf{30.1} & 12.1 \\

PETRv1$\dagger$~\cite{liu2022petr} & V2-99~\cite{lee2019energy} & 50.4 & 44.1 & 59.3 & 24.9 & 38.3 & 80.8 & 13.2 \\

PETRv2$\dagger$~\cite{liu2022petrv2} & V2-99~\cite{lee2019energy} & 58.2 & 49.0 & 56.1 & 24.3 & 36.1 & 34.3 & \textbf{12.0} \\

\Xhline{2.0\arrayrulewidth}
BEVFormer~\cite{li2022bevformer} & V2-99~\cite{lee2019energy} & 56.9 & 48.1 & 58.2 & 25.6 & 37.5 & 37.8 & 12.6 \\

\rowcolor{gray!15} \textbf{VectorFormer} & V2-99~\cite{lee2019energy} & \textbf{58.3} \textbf{\scriptsize{(\texttt{+}1.4)}} & \textbf{49.2} \textbf{\scriptsize{(\texttt{+}1.1)}} & 56.3 & 25.2 & \textbf{35.2} & 33.8 & 12.7 \\

\Xhline{2.0\arrayrulewidth}

\end{tabular}
}
\end{center}

\caption{
  \textbf{Comparison of recent works on nuScenes detection~\cite{caesar2020nuscenes} test set.} Methods trained with CBGS~\cite{zhuClassbalancedGroupingSampling2019a} are indicated with $\dagger$. The best results among methods are in bold.
}
\label{tab:all_metrics_test_set}

\end{table}

\begin{table}[t!]
\begin{center}
\centering
\resizebox{\textwidth}{!}{
\setlength{\tabcolsep}{0.3em}
\renewcommand{\arraystretch}{1.05}
\begin{tabular}{l | l | c c | c c c c c}
\Xhline{2.0\arrayrulewidth}
Method & Backbone & NDS~$\uparrow$ & mAP~$\uparrow$ & mATE~$\downarrow$ & mASE~$\downarrow$ & mAOE~$\downarrow$ & mAVE~$\downarrow$ & mAAE~$\downarrow$ \\
\Xhline{2.0\arrayrulewidth}

BEVDet$\dagger$~\cite{huang2021bevdet} & Swin-T~\cite{liu2021swin} & 41.7 & 34.9 & 63.7 & 26.9 & 49.0 & 91.4 & 26.8 \\

DETR3D~\cite{wang2022detr3d} & ResNet-101 & 42.5 & 34.6 & 77.3 & 26.8 & 38.3 & 84.2 & 21.6 \\

DETR4D~\cite{luo2022detr4d} & ResNet-101 & 50.9 & 42.2 & 68.8 & 26.9 & 38.8 & 49.6 & \textbf{18.4} \\

PETRv1$\dagger$~\cite{liu2022petr} & ResNet-101 & 44.2 & 37.0 & 71.1 & 26.7 & 38.3 & 86.5 & 20.1 \\

PETRv2$\dagger$~\cite{liu2022petrv2} & ResNet-101 & 52.4 & 42.1 & 68.1 & 26.7 & 35.7 & 37.7 & 18.6 \\

3DPPE~\cite{shu20233dppe}$\dagger$ & ResNet-101 & 45.8 & 39.1 & 67.4 & 28.2 &  39.5 & 83.0 & 19.1 \\

OCBEV~\cite{qi2023ocbev} & ResNet-101 & 53.2 & 41.7 & 62.9 & 27.3 & \textbf{33.9} & 34.2 & 18.7 \\

AeDet$\S$~\cite{feng2023aedet} & ResNet-101 & 50.6 & 39.4 & \textbf{60.9} & \textbf{26.6} & 41.2 & 42.0 & 20.1 \\

\Xhline{2.0\arrayrulewidth}
BEVFormer-S~\cite{li2022bevformer} & ResNet-101 & 47.9 & 37.0 & 72.1 & 28.0 & 40.7 & 43.6 & 22.0 \\

\rowcolor{gray!15} \textbf{VectorFormer-S} & ResNet-101 & 51.0 \textbf{\scriptsize{(\texttt{+}3.1)}} & 40.5 \textbf{\scriptsize{(\texttt{+}3.5)}} & 67.6 & 27.3 & 38.9 & 39.9 & 19.2 \\

\Xhline{2.0\arrayrulewidth}
DFA3D-S~\cite{li2023dfa3d} & ResNet-101 & 50.1 & 40.1 & 72.1 & 27.9 & 41.1 & 39.1 & 19.6 \\

\rowcolor{gray!15}\textbf{VectorFormer-DFA3D-S} & ResNet-101 & 51.4 \textbf{\scriptsize{(\texttt{+}1.3)}} & 40.4 \textbf{\scriptsize{(\texttt{+}0.3)}} & 68.1 & 27.4 & 35.4 & 36.4 & 20.5 \\

\Xhline{2.0\arrayrulewidth}
BEVFormer-B~\cite{li2022bevformer} & ResNet-101 & 51.7 & 41.6 & 67.3 & 27.4 & 37.2 & 39.4 & 19.8 \\

\rowcolor{gray!15} \textbf{VectorFormer-B} & ResNet-101 & 53.2 \textbf{\scriptsize{(\texttt{+}1.5)}} & 42.5 \textbf{\scriptsize{(\texttt{+}0.9)}} & 64.3 & 27.5 & 35.2 & 34.4 & 18.8 \\

\Xhline{2.0\arrayrulewidth}

DFA3D-B~\cite{li2023dfa3d} & ResNet-101 & 53.1 & 43.0 & 65.4 & 27.1 & 37.4 & 34.1 & 20.5 \\

\rowcolor{gray!15} \textbf{VectorFormer-DFA3D-B} & ResNet-101 & \textbf{54.0 \scriptsize{(\texttt{+}0.9)}} & \textbf{43.7 \scriptsize{(\texttt{+}0.7)}} & 64.3 & 27.0 & 36.3 & \textbf{32.4} & 18.6 \\
\Xhline{2.0\arrayrulewidth}

\Xhline{2.0\arrayrulewidth}
\end{tabular}
}
\end{center}

\caption{
  \textbf{Comparison of recent works on nuScenes detection~\cite{caesar2020nuscenes} validation set.} Methods trained with CBGS~\cite{zhuClassbalancedGroupingSampling2019a} are indicated with $\dagger$. Results reproduced for fair comparison are indicated as $\S$. The best results among methods are in bold.
}
\label{tab:all_metrics_val}

\end{table}

\section{Experiments}
\subsection{Experimental Setup}
\noindent
\textbf{nuScenes Dataset}
We conduct our experiments on the nuScenes~\cite{caesar2020nuscenes} dataset with 1000 scenes. Each scene sample contains six-view RGB images. We evaluate the 3D detectors with the metrics of NDS, mAP, mATE, mASE, mAOE, mAVE, and mAAE as the existing works~\cite{li2022bevformer,li2023dfa3d}.

\noindent
\textbf{Waymo Open Dataset} We evaluate our method on the Waymo dataset~\cite{sun2020scalability}, which provides five-view images covering 252$^{\circ}$ horizontal FOV. We follow the settings as~\cite{li2022bevformer,li2022delving} to experiment on the same subset of the training split for a fair comparison and evaluate with the metrics of LET-3D-mAP, LET-3D-mAH, and LET-3D-mAL~\cite{hung2022let}.

\begin{table}[t!]
\begin{center}
\centering
\resizebox{\textwidth}{!}{
\setlength{\tabcolsep}{0.3em}
\renewcommand{\arraystretch}{1.05}
\begin{tabular}{l | l | c c | c c c c c}
\Xhline{2.0\arrayrulewidth}
Method & Backbone & NDS~$\uparrow$ & mAP~$\uparrow$ & mATE~$\downarrow$ & mASE~$\downarrow$ & mAOE~$\downarrow$ & mAVE~$\downarrow$ & mAAE~$\downarrow$ \\
\Xhline{2.0\arrayrulewidth}

BEVDet4D~\cite{huang2022bevdet4d} & Swin-B~\cite{liu2021swin} & 51.5 & 39.6 & 61.9 & 26.0 & 36.1 & 39.9 & 18.9 \\

PETRv2~\cite{liu2022petrv2} & V2-99~\cite{lee2019energy} & 50.3 & 41.0 & 72.3 & 26.9 &  45.3  & 38.9 & 19.3 \\

SA-BEV$^\dagger$~\cite{zhang2023sa} & ConvNeXt-B~\cite{liu2022convnet} & 57.9 & 47.9 & - & - & - & - & - \\

StreamPETR~\cite{Wang_2023_ICCV} & V2-99~\cite{lee2019energy} & 57.1 & 48.2 & 61.0 & \textbf{25.6} & 37.5  & \textbf{26.3} & 19.4 \\

SparseBEV$^\star$~\cite{liu2023sparsebev} & V2-99~\cite{lee2019energy} & 57.9 & 49.4 & - & - & -  & - & - \\

\rowcolor{gray!15} \textbf{VectorFormer} & V2-99~\cite{lee2019energy} & \textbf{60.5} & \textbf{51.7} & \textbf{57.0} & 26.6 & \textbf{22.8} & 28.7 & \textbf{18.7} \\
\Xhline{2.0\arrayrulewidth}
\end{tabular}
}
\end{center}
\caption{
  \textbf{Comparison of recent works trained with no more than 24 epochs on nuScenes detection~\cite{caesar2020nuscenes} validation set using large backbone.} $^\dagger$ indicates methods trained with CBGS~\cite{zhuClassbalancedGroupingSampling2019a}. $^\star$ indicates a method with dual branches design. The best results among methods are in bold.
}

\label{tab:all_metrics_val_large}
\end{table}

\begin{table}[t!]
\begin{center}
\centering
\resizebox{\textwidth}{!}{
\setlength{\tabcolsep}{0.3em}
\renewcommand{\arraystretch}{1.05}
\begin{tabular}{l | r | c | c c| c | c}
\Xhline{2.0\arrayrulewidth}
Method & BEV Dim. & $x$- and $y$-axis Vector Dim. & NDS~$\uparrow$ & mAP~$\uparrow$ & FPS & Mem. \small{(GB)} \\
\Xhline{2.0\arrayrulewidth}
 
BEVFormer~\cite{li2022bevformer} 
 & $200 \times 200$ & - & 51.7 & 41.6 & 3.9 & 4.70 \\
 
\textbf{Ours}
 & $150 \times 150$ & $1\times200$ \& $200\times1$  & 52.8 & 41.8 & 3.9 & 4.78 \\
 
\Xhline{2.0\arrayrulewidth}

BEVFormer~\cite{li2022bevformer} 
 & $300 \times 300$ & - & 51.9 & 41.4 & 3.2 & 5.58 \\
 
\textbf{Ours}
 & $200 \times 200$ & $1\times300$ \& $300\times1$  & 53.0 & 42.1 & 3.5 & 4.83 \\

\Xhline{2.0\arrayrulewidth}

BEVFormer~\cite{li2022bevformer} 
 & $450 \times 450$ & - & - & - & 2.3 & 11.21 \\
 \textbf{Ours}
 & $200 \times 200$ & $1\times450$ \& $450\times1$  & 53.2 & 42.5 & 3.4 & 4.85 \\
 
\Xhline{2.0\arrayrulewidth}

\end{tabular}
}
\end{center}

\caption{
\textbf{Effectiveness and efficiency comparisons between BEVFormer~\cite{li2022bevformer} and our proposed VectorFormer.} BEVFormer with a BEV dimension of $450\times450$ could not be trained with NVIDIA A100 40G GPUs because of the expensive memory consumption. Frames-per-second (FPS) are tested with NVIDIA RTX4090 GPU.
}
\label{tab:resolution}

\end{table}

\noindent
\textbf{Implementation Details}
For the nuScenes dataset~\cite{caesar2020nuscenes}, we followed the typical practice as~\cite{li2022bevformer,wang2022detr3d} and adopted the ResNet101-DCN~\cite{he2016deep,dai2017deformable} with pretrained checkpoint from FCOS3D~\cite{wang2021fcos3d} as image backbone to process the images with the resolution of $1600\times900$. Similarly, BEV queries are initialized with the size of $200\times 200$ for the base setting and $150\times 150$ for the small setting. The vector queries are initialized with a length of $450$. The numbers of encoding/decoding layers and used history frames are maintained the same as BEVFormer~\cite{li2022bevformer} under similar model settings. To scale up our method, we experiment with the large image backbone of V2-99~\cite{lee2019energy}  with the pretrained DD3D~\cite{park2021pseudo} checkpoint. On the Waymo dataset~\cite{sun2020scalability}, we followed the practice as~\cite{li2022bevformer,li2022delving} to make a fair comparison. All models are trained with 24 epochs with a learning rate of $2\times10^{-4}$ as the baseline~\cite{li2022bevformer}. 

\begin{figure}[t!]
\centering
\includegraphics[width=1.0\textwidth]{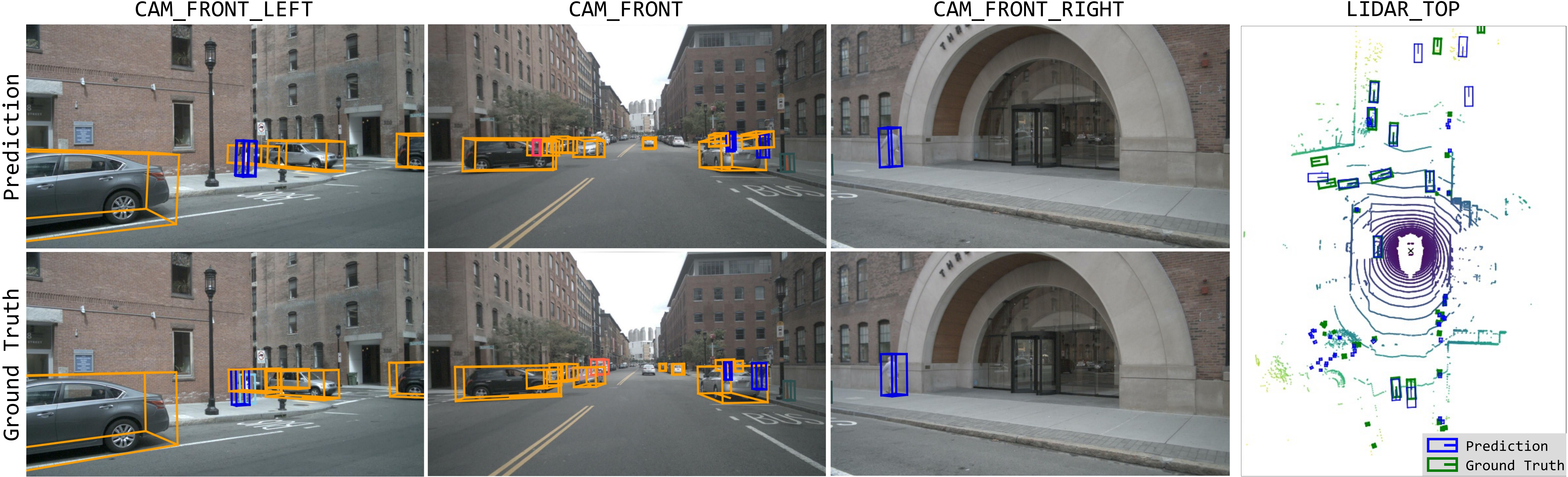}

\caption{\textbf{Visualization results of VectorFormer on nuScenes~\cite{caesar2020nuscenes} validation set.} Detection predictions with ground truth in multi-view camera images are shown on the left and in bird's-eye-view is shown on the right.}
\label{fig:visualDet}

\end{figure}

\subsection{Main Results}
We demonstrate the 3D detection results of our proposed VectorFormer on the nuScenes~\cite{caesar2020nuscenes} test set in Tab.~\ref{tab:all_metrics_test_set} and validation set in Tab.~\ref{tab:all_metrics_val} and Tab.~\ref{tab:all_metrics_val_large}. We also evaluate our method on the Waymo dataset~\cite{sun2020scalability} in Tab.~\ref{tab:waymo}.

\noindent
\textbf{nuScenes Dataset} 
Without using any bells and whistles, we achieve the best performance and present with remarkable improvements against BEVFormer~\cite{li2022bevformer}, with 1.4 points higher on NDS (58.3\% vs. 56.9\%) and 1.1 points higher on mAP (49.2\% vs. 48.1\%) on nuScenes~\cite{caesar2020nuscenes} test set, as shown in Tab.~\ref{tab:all_metrics_test_set}.

We also compare our small and base settings with previous state-of-the-art methods on the nuScenes~\cite{caesar2020nuscenes} validation set in Tab.~\ref{tab:all_metrics_val}. Besides, we extend our proposed vector representation to a recent work of DFA3D~\cite{li2023dfa3d}. Concretely, VectorFormer-S outperforms the baseline BEVFormer-S~\cite{li2022bevformer} with 3.1 points in NDS and 3.5 points in mAP. When extending our method to DFA3D~\cite{li2023dfa3d}, our VectorFormer-DFA3D-S consistently achieves a leap upon DFA3D-S~\cite{li2023dfa3d} with 1.3 points in NDS and 0.3 points in mAP. Regarding the base setting, our VectorFormer-B presents improvements of 1.5 points in NDS and 0.9 points in mAP compared to the BEVFormer-B~\cite{li2022bevformer}. When extended to the DFA3D-B, we fine-tuned from the DFA3D's~\cite{li2023dfa3d} pretrained checkpoint by freezing the image backbone to save the GPU memory cost. It is observed that we can further boost the performance by achieving a higher NDS of 54.0\%. It is worth mentioning that our proposed VectorFormer-S and VectorFormer-B, which do not utilize depth information, still present a superior NDS against DFA3D-S~\cite{li2023dfa3d} (51.0\% vs. 50.1\%) and a comparable NDS to DFA3D-B~\cite{li2023dfa3d} (53.2\% vs. 53.1\%). When scaling up the method with a larger image backbone, we also demonstrate superior results upon the recent SOTAs, as shown in Tab.~\ref{tab:all_metrics_val_large}.

Overall, we achieve state-of-the-art performance, and our proposed vector query representation can significantly improve query-BEV-based methods.

\begin{table}[t!]

\begin{center}
\centering
\resizebox{0.8\linewidth}{!}{
\setlength{\tabcolsep}{0.3em}
\renewcommand{\arraystretch}{1.05}
\begin{tabular}{l|c|c|c}

\Xhline{2.0\arrayrulewidth}
Methods & LET-mAPL$\uparrow$ & LET-mAPH$\uparrow$ & LET-mAP$\uparrow$ \\
\Xhline{2.0\arrayrulewidth}

MV-FCOS3D++~\cite{wang2022mv}
& \textbf{37.9} & 48.8 & 52.2 \\

BEVFormer~\cite{li2022bevformer}
& 35.0 & 47.1 & 51.0 \\

\rowcolor{gray!15}\textbf{VectorFormer}
& 36.8 & \textbf{49.1} & \textbf{53.2} \\

\Xhline{2.0\arrayrulewidth}
\end{tabular}
 }
\end{center}

\caption{
Comparison on Waymo validation set.
}
\label{tab:waymo}

\end{table}

\noindent
\textbf{Representation Dimension and Computation Overhead}
In Tab.~\ref{tab:resolution}, we illustrate the performance and computation overhead details of Fig.~\ref{fig:motivation}. When growing the BEV resolution, the traditional framework shows performance vanishing and leads to computational cost explosions, failing to take advantage of the increases in the representation dimension. In comparison, our method with the vector representation enjoys finer BEV granularity construction with $O(N)$ complexity. Comparing under a similar level of BEV resolution, we can speed up the FPS by up to 47.8\% and save the GPU memory consumption by up to 56.7\% while achieving better performance.

\noindent
\textbf{Qualitative Results}
As shown in Fig.~\ref{fig:visualDet}, our proposed VectorFormer presents a conspicuous detection performance. Even for objects with severe occlusion or located further apart, the model can still produce satisfactory bounding boxes.

\noindent
\textbf{Waymo Dataset}
In Tab~\ref{tab:waymo}, we further evaluate our method on the dataset of Waymo~\cite{sun2020scalability}. Our proposed VectorFormer consistently achieves better results towards the baselines.

\begin{table}[t!]
\begin{center}
\centering
\resizebox{\textwidth}{!}{
    \setlength{\tabcolsep}{0.3em}
      \renewcommand{\arraystretch}{1.05}
\begin{tabular}{ccc|cc|ccccc}

\Xhline{2.0\arrayrulewidth}
Heatmap & Spatial & Temporal & NDS~$\uparrow$ & mAP~$\uparrow$ & mATE~$\downarrow$ & mASE~$\downarrow$ & mAOE~$\downarrow$ & mAVE~$\downarrow$ & mAAE~$\downarrow$ \\
\Xhline{2.0\arrayrulewidth}
- & - & - 
& 48.4 & 38.1 & 72.8 & 27.8 & 39.9 & 46.2 & \textbf{19.7} \\ 

\cmark & - & - 
& 48.9 & 38.8 & 72.1 & 27.8 & 41.3 & 43.5 & 20.0 \\ 

\cmark & \cmark & - 
& 49.5 & 39.3  & \textbf{69.6} & 27.6 & 41.4 & \textbf{42.6} & 20.6 \\ 

\Xhline{2.0\arrayrulewidth}
\cmark & \cmark & \cmark 
& \textbf{49.7} & \textbf{39.8} & 70.9 & \textbf{27.4} & \textbf{38.8} & 44.5 & 20.0 \\

\Xhline{2.0\arrayrulewidth}
\end{tabular}
}
\end{center}

\caption{
\textbf{The ablation studies of different components in our proposed VectorFormer.} Heatmap indicates using heatmap supervision for the BEV features map, Spatial indicates incorporating our proposed vector representations for spatial modeling, and Temporal indicates conducting temporal modeling for the vector representation.}
\label{tab:components}
\end{table}

\begin{table}[t!]
   \begin{center}
   \begin{minipage}{.5\linewidth}
   \begin{center}
      \resizebox{\textwidth}{!}{
      \setlength{\tabcolsep}{0.3em}
      \renewcommand{\arraystretch}{1.2}
      \begin{tabular}{c|ccccc}
        \Xhline{2.0\arrayrulewidth}
         Vector Combin. & NDS & mAP & mATE & mASE & mAOE \\
        \Xhline{2.0\arrayrulewidth}
            Mult. 
            & 49.4 & 39.4 & \textbf{70.2} & 27.6 & 42.3 \\
            Add.
            & \textbf{49.7} & \textbf{39.8} & 70.9 & \textbf{27.4} & \textbf{38.8} \\ 
        \Xhline{2.0\arrayrulewidth}
   \end{tabular}
   }
   \caption{The ablation study on the combination of vector query.}
   \label{tab:multandadd}
   \end{center}
   \end{minipage}
   \begin{minipage}{.45\linewidth}
   \begin{center}
   \resizebox{\textwidth}{!}{
   \setlength{\tabcolsep}{0.3em}
   \begin{tabular}{c|ccccc}

        \Xhline{2.0\arrayrulewidth}
        Top-$k$ Proposals & NDS & mAP & mATE & mASE & mAOE \\
        \Xhline{2.0\arrayrulewidth}
        2 
        & 49.5 & 39.6 & 69.6 & 27.9 & 41.7 \\
        3
        & 49.7 & 39.8 & 70.9 & 27.4 & 38.8 \\ 
        4
        & 50.0 & 40.0 & 69.0 & 27.3 & 40.2 \\

        \Xhline{2.0\arrayrulewidth}
   \end{tabular}
   }
   \caption{The ablation study on the number of proposal positions.}
   \label{tab:topk}
   \end{center}
   \end{minipage}
   \end{center}
\end{table}
\subsection{Ablation Study}
\label{sec4.3}
All the experiments in this section are conducted with VectorFormer-S by default with 12 training epochs. 

\noindent
\textbf{Effect of Model Components} As illustrated in Tab.~\ref{tab:components}, we present the ablation studies on the effect of different component designs in VectorFormer. It is demonstrated that our proposed framework design for learning high-resolution vector representation is effective, and its components work in synergy to uplift the detection performance with 1.3 points in NDS and 1.7 points in mAP.

\noindent
\textbf{Effect of Vector Query Combination} We factorize the high-resolution (HR) BEV features into $x$ and $y$ vector features. We conduct ablation studies on the operators of multiply and addition for combining these factorized vectors to build HR BEV features, as illustrated in Tab.~\ref{tab:multandadd}. It is shown that combining the vectors to build HR BEV features by addition leads to better performance.

\noindent
\textbf{Effect of Foreground Proposal Number} As shown in Tab.~\ref{tab:topk}, we conduct ablation studies on the number of foreground proposal positions along the $x$ and $y$ directions. Overall, the performance will improve as we increase the number of proposals, and we choose to use three proposals in each direction in practice.

\noindent
\textbf{Effect of Offset Prediction} We represent the foreground proposal position with HR BEV queries and further adaptively deform the positions with predicted offsets. This practice improves the mAP with 0.4 points, as illustrated in the first and the fourth rows of Tab.~\ref{tab:offsetpredict}.

\noindent
\textbf{Effect of LR-HR Fusion} The evenly sampled LR BEV features and sparse HR BEV features independently interact with the image features. Therefore, we apply fusion between LR and HR features before (Eqn.~\ref{eqn:6}) and after (Eqn.~\ref{eqn:8}) passing the SCA module, and it results in improvements in overall performance, as shown in the second and the fourth rows of Tab.~\ref{tab:offsetpredict}.

\noindent
\textbf{Effect of Positional Embeddings} The sparse HR query located at $(x,y)$ is composited according to the Eqn.~\ref{eqn:1}. When aggregating the sparse HR query at $(x,y)$ to $\mathbf{V^X}$ or $\mathbf{V^Y}$, we use their positional embeddings $\mathbf{PE^Y}$ or $\mathbf{PE^X}$ that correspond to the dimension about to collapse to remedy the information loss (Eqn.~\ref{eqn:9}). Comparing the third and the fourth rows In Tab.~\ref{tab:offsetpredict}, it is shown that the usage of positional embeddings as the attention keys contributes to the overall performance improvements.

\begin{table}[t!]
\begin{center}
\centering
\resizebox{\textwidth}{!}{
\setlength{\tabcolsep}{0.3em}
      \renewcommand{\arraystretch}{1.05}
\begin{tabular}{ccc|ccccc}

\Xhline{2.0\arrayrulewidth}
 Offset Pred. & LR-HR Fusion & Pos. Emb. & NDS & mAP & mATE & mASE & mAOE \\
\Xhline{2.0\arrayrulewidth}

- & \cmark & \cmark
& \textbf{49.7} & 39.2 & 70.0 & 27.7 & 39.7 \\

\cmark & - & \cmark
& 49.6 & 39.3 & \textbf{69.0} & \textbf{27.4} & 41.3 \\

\cmark & \cmark & -
& 49.3 & 39.1 & 70.6 & 27.7 & 41.6 \\

\Xhline{2.0\arrayrulewidth}
\cmark & \cmark & \cmark
& \textbf{49.7} & \textbf{39.8} & 70.9 & \textbf{27.4} & \textbf{38.8} \\

\Xhline{2.0\arrayrulewidth}
\end{tabular}
}
\end{center}

\caption{
The ablation study on offsetting the proposal positions, conducting fusion on sparse HR BEV queries with the LR BEV queries, and using positional embeddings in the Vector query gathering module.
}
\label{tab:offsetpredict}

\end{table}

\section{Conclusion}
In this paper, we have presented a camera-based 3D object detector, VectorFormer, accompanied by a novel representation of Vector query. Addressing the limitations of traditional BEV queries, which incur substantial computational costs and memory usage as spatial resolution increases, we propose to utilize a more lightweight representation of vector query, focusing on learning finer-grained representations for the crucial regions through our designed vector query scattering and gathering modules. The vector queries compact richer spatial and temporal priors with low complexity of time and memory, which are further used to enhance the decoder to produce more accurate predictions. The extensive experiments demonstrate that our proposed VectorFormer achieves state-of-the-art performance, and the designed framework is generalizable, which can consistently leap the performance of the query-BEV methods.

\clearpage  

\section*{Acknowledgements}
The authors are thankful for the financial support from the Hetao Shenzhen-HongKong Science and Technology Innovation Cooperation Zone (HZQB-KCZYZ-2021055), this work was also supported by Shenzhen Deeproute.ai Co., Ltd (HZQB-KCZYZ-2021055).
%
%
\bibliographystyle{splncs04}
\bibliography{main}
\end{document}